\pdfoutput=1
\documentclass[conference]{IEEEtran}
\IEEEoverridecommandlockouts

\usepackage{cite}
\usepackage{amsmath,amssymb,amsfonts}
\usepackage{algorithmic}
\usepackage{graphicx}
\usepackage{textcomp}
\usepackage{xcolor}
\usepackage{url}
\def\BibTeX{{\rm B\kern-.05em{\sc i\kern-.025em b}\kern-.08em
    T\kern-.1667em\lower.7ex\hbox{E}\kern-.125emX}}
\begin{document}

\title{Embodied Zero-Shot Task Planning via Hardware-Aware LLM–VLM Integration for Grounded Robotic Manipulation\\
}

\author{\IEEEauthorblockN{Lonnie Tankersley}
\IEEEauthorblockA{\textit{dept. of Aerospace Engineering} \\
\textit{Mississippi State University}\\
Starkville, MS \\
lmt413@msstate.edu }
\and
\IEEEauthorblockN{Noorbakhsh Amiri Golilarz}
\IEEEauthorblockA{\textit{dept. of Computer Science} \\
\textit{The University of Alabama}\\
Tuscaloosa, AL \\
namirigolilarz@ua.edu}
\and
\IEEEauthorblockN{Soroush Korivand}
\IEEEauthorblockA{\textit{dept. of Mechanical Engineering} \\
\textit{Mississippi State University}\\
Starkville, MS \\
skorivand@me.msstate.edu}
\thanks{\copyright~2026 IEEE. Personal use of this material is permitted. Permission from IEEE must be obtained for all other uses, in any current or future media, including reprinting/republishing this material for advertising or promotional purposes, creating new collective works, for resale or redistribution to servers or lists, or reuse of any copyrighted component of this work in other works.}
}

\maketitle

\begin{abstract}
Large language models can translate natural-language instructions into plausible manipulation plans, but robotic execution requires physically grounded decisions that respect robot embodiment, workspace geometry, reachability, collision constraints, and grasp feasibility. This paper presents a hardware-aware large language model–vision-language model (LLM–VLM) manipulation framework for a simulated Franka Research 3 robot in NVIDIA Isaac Sim. The framework separates symbolic reasoning from metric execution: an API-hosted LLM planner receives a structured scene summary and generates only symbolic task information, including the object to manipulate, target relation, approach preference, and objects to avoid. A local Qwen3-VL 8B VLM verifier supports semantic confirmation of ambiguous objects and open-container surrogates. Both the LLM planner and the VLM verifier are prohibited from generating robot coordinates, grasp poses, keepout dimensions, gripper commands, trajectory waypoints, or success labels. Instead, these decisions are handled by deterministic grounding and execution modules that use the robot model, scene geometry, MoveIt 2 collision checking, inverse kinematics, grasp validation, release verification, and final visual assessment. Execution feedback is returned only as symbolic context for replanning. The framework is evaluated through simulated digital-twin trials involving pickup, occluded-source access, keepout reasoning, relation placement, and sequential manipulation tasks.
\end{abstract}


\section{INTRODUCTION}

Large Language Models (LLMs), Vision-Language Models (VLMs), and Vision-Language-Action (VLA) models are rapidly changing the way robotic systems interpret human instructions and connect language to action. Language-conditioned manipulation systems have shown that pretrained models can rank candidate robot skills using learned affordance scores \cite{SayCan}, generate executable programs over predefined robot application programming interfaces (APIs) \cite{CodeAsPolicies}, decompose high-level commands into admissible action sequences \cite{ZeroShotPlanner-Huang,progpromt}, and support replanning through feedback from the environment or the human user \cite{InnerMonologue}. Other approaches have used language models to define intermediate objectives, such as rewards, costs, or value maps, that guide downstream motion planners and controllers rather than directly producing robot commands \cite{VoxPoser}. Together, these works establish an important design pattern: foundation models are useful for interpreting intent and proposing task structure, but physical execution must still be grounded in the robot, the scene, and the available control interface.

In parallel, embodied multimodal and VLA models have moved toward more direct integration of visual observations, natural-language instructions, robot state, and action generation. Large embodied models have demonstrated that visual tokens, language tokens, and robot state can be integrated for embodied reasoning \cite{Palm-E}, while VLA models have shown that pretrained vision-language representations can be adapted to robotic control through robot trajectory data and action representations compatible with language-model backbones \cite{RT-2}. Broader robot-learning datasets and multi-embodiment training efforts further suggest that generalist policies can benefit from demonstrations across diverse tasks, environments, and platforms \cite{RT-1,Open-X}. More recent open-source and continuous-action VLA models continue this trend by improving accessibility, fine-tuning efficiency, and control frequency across manipulation settings \cite{openVLA,VLA-FlowModel}. These systems represent an important step toward general-purpose robot control, but they also make auditability more difficult because perception, language grounding, and action generation are often tightly coupled. When a manipulation attempt fails, it may be unclear whether the cause was incorrect visual interpretation, ambiguous language grounding, invalid action generation, hardware incompatibility, or execution dynamics.

This distinction is critical because many decisions that appear semantic at the language level are physically constrained at the execution level. A model may correctly identify that a cube should be picked, that a container should receive an object, or that an obstacle should be avoided, while still producing or implying an invalid grasp, unreachable pose, unsafe approach direction, or geometrically incorrect placement. These errors arise because foundation models do not inherently possess a calibrated model of the robot's kinematics, gripper geometry, workspace limits, collision environment, or current depth-derived scene state. A placement target that is semantically reasonable may be outside the robot's reachable workspace. A grasp that is linguistically plausible may exceed the gripper aperture or approach the object from an occluded side. A keepout region may be necessary around a nearby object, but its size and shape must be derived from the physical scene and end-effector geometry, not from language-model inference. Therefore, the main challenge is not simply whether an LLM or VLM can generate a reasonable plan, but whether the system architecture prevents physically ungrounded model outputs from becoming robot commands.

The grounding of language-based plans also depends on perception systems capable of linking natural-language object references to physical scene elements. Open-vocabulary detection methods allow objects to be detected from flexible category names and referring expressions rather than from a fixed closed vocabulary \cite{Cheng_2024_CVPR,DINO}, while promptable segmentation methods improve mask generation and refinement in cluttered or partially occluded scenes \cite{Sam2}. Beyond detection and segmentation, open-vocabulary maps, language-embedded 3D scene representations, and hierarchical scene graphs make environments more queryable for language-conditioned planning \cite{Open-vocabulary-Queryable}. These methods provide useful semantic access to the scene, but they primarily address what objects are present and where they are located. They do not, by themselves, determine whether a specific robot can safely manipulate a detected object, whether the object is compatible with the available gripper, or whether occlusion requires a top-down approach, side approach, re-observation, or task rejection.

This problem has been recognized in several forms. Surveys of foundation models in robotics identify hallucination, weak uncertainty estimation, and limited real-time safety guarantees as persistent barriers to deployment \cite{firoozi2025foundation}. Empirical work has also shown that adversarial prompts can cause LLM-controlled robots to execute unsafe actions when textual outputs are allowed to influence physical behavior without a robust authority boundary \cite{Jailbreaking}. Existing systems partially address this issue through affordance scoring, prompt constraints, executable API definitions, human feedback, learned feasibility models, or post-hoc filtering \cite{SayCan,progpromt,RobotAsk4Help}. While these mechanisms improve reliability, they do not fully resolve the architectural question of which decisions a foundation model is allowed to make. In many pipelines, model outputs can still influence metric execution decisions such as object coordinates, grasp poses, trajectory waypoints, spatial tolerances, or success labels. As a result, physical validity is often treated as a downstream correction problem rather than as a formally enforced separation of authority.

This paper argues that a formal separation of authority is essential for auditable foundation-model-based manipulation. Rather than treating an LLM or VLM as a controller whose outputs are filtered after generation, the proposed framework treats foundation models as bounded semantic hypothesis generators. The model may propose symbolic task structure, including the source object, target object, placement relation, approach preference, and symbolic obstacle identifiers, but it cannot emit metric robot commands such as coordinates, joint angles, gripper widths, keepout dimensions, trajectory waypoints, or success labels. Those physically consequential decisions are assigned to deterministic robot-owned modules responsible for grounding, motion feasibility, execution, and verification. This design shifts foundation models from physical authority to semantic assistance, while making failures attributable to specific stages such as schema validation, object grounding, motion execution, or final visual verification rather than to an opaque end-to-end model. Its contribution is therefore not a new general-purpose VLA policy or a claim of mature manipulation performance, but an auditable zero-shot manipulation architecture in which success and failure can be inspected through explicit validation boundaries. The implementation, evaluation artifacts, and supporting materials are available in the project repository.\footnote{\url{https://github.com/Ltankersley15/SK-Zero-Shot-Research}}
Accordingly, the specific contributions of this paper are as follows:

\begin{enumerate}
    \item \textbf{A bounded-authority LLM--VLM manipulation framework.} The framework constrains foundation-model outputs to typed symbolic and semantic fields, while deterministic modules retain authority over pose selection, collision-aware grounding, gripper control, motion execution, and outcome verification.

    \item \textbf{Hardware-aware deterministic grounding for symbolic plans.} The system grounds symbolic outputs using the robot kinematics, gripper geometry, workspace constraints, depth-based object state, and occlusion-aware access annotations, enabling approach, grasp, avoidance, and placement decisions without giving foundation models authority over physical geometry.

    \item \textbf{Failure-localized evaluation of zero-shot manipulation trials.} The evaluation decomposes each trial into schema validation, grounding feasibility, motion execution, and visual verification, enabling auditable failure diagnosis beyond binary task success.

    \item \textbf{Boundary-preserving replanning from execution feedback.} The executor returns force, contact, grasp, or placement failures as symbolic feedback for replanning, while each revised plan remains subject to the same schema constraints and authority boundary.
\end{enumerate}

\section{MATERIALS AND METHODS}

\subsection{System Overview}

\begin{figure*}[t]
    \centering
    \includegraphics[width=0.8\textwidth]{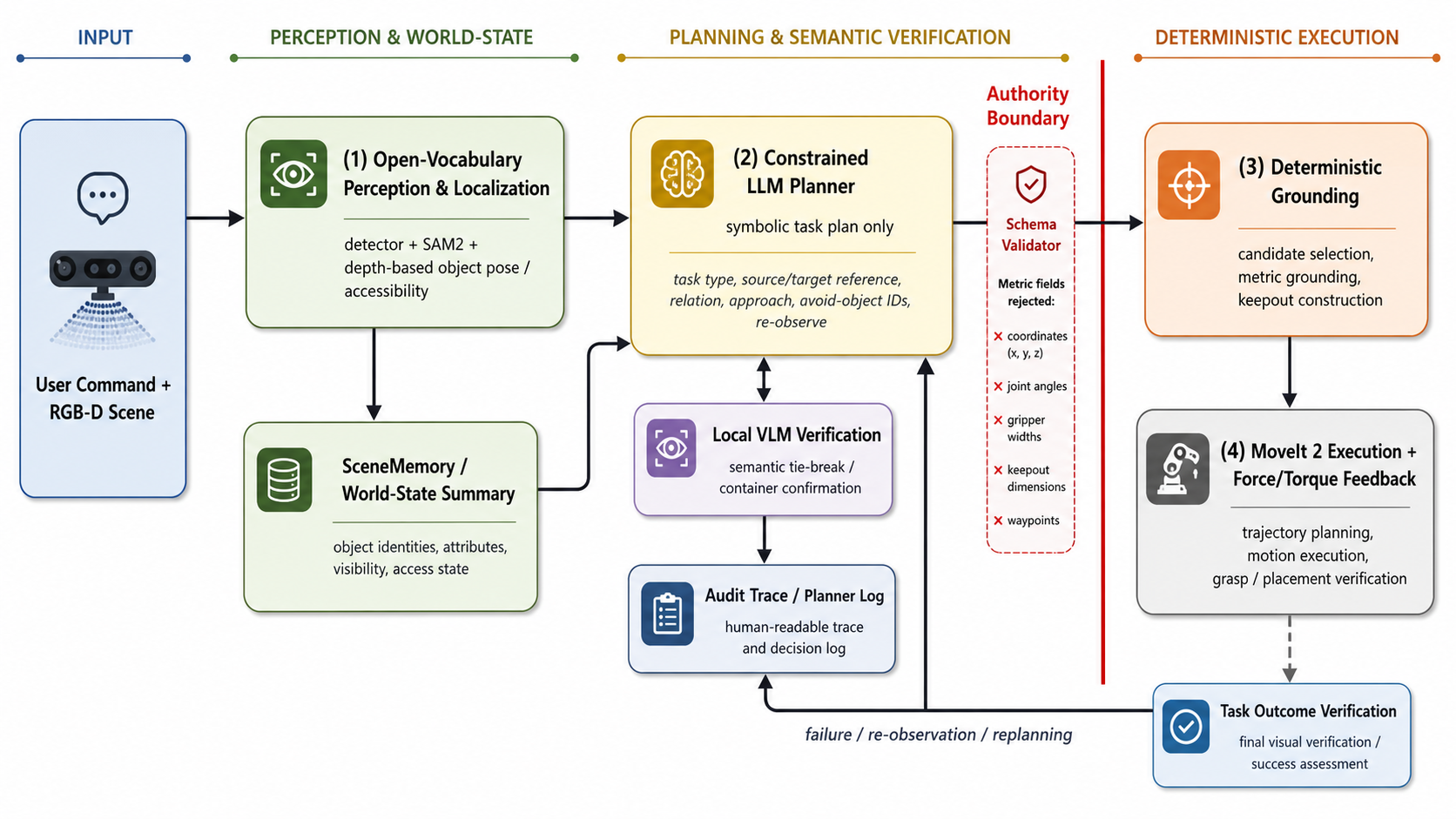}
    \vspace{-0.5cm}
    \caption{The proposed manipulation framework converts a user command and RGB-D scene into a structured world state, then performs constrained symbolic planning, authority-boundary validation, deterministic grounding, MoveIt~2 execution, and feedback-based replanning.}
    \label{fig:system_architecture}
    \vspace{-0.5cm}
\end{figure*}
The proposed framework follows a perception--planning--grounding--execution pipeline, as shown in Fig.~\ref{fig:system_architecture}. 

Each trial begins with a natural-language command and an RGB-D observation of the tabletop scene, from which the perception layer constructs a structured world-state representation called \textit{SceneMemory}. This representation stores object hypotheses, semantic labels, visual masks, approximate pose proxies, visibility states, and symbolic access annotations. The language planner operates only on the structured scene summary and returns a symbolic task plan. A schema validator then enforces the authority boundary, preventing metric robot commands from entering the execution layer.

\subsection{Simulation Platform and Scene Representation}

\begin{figure}[t]
    \centering
    \includegraphics[width=0.65\linewidth]{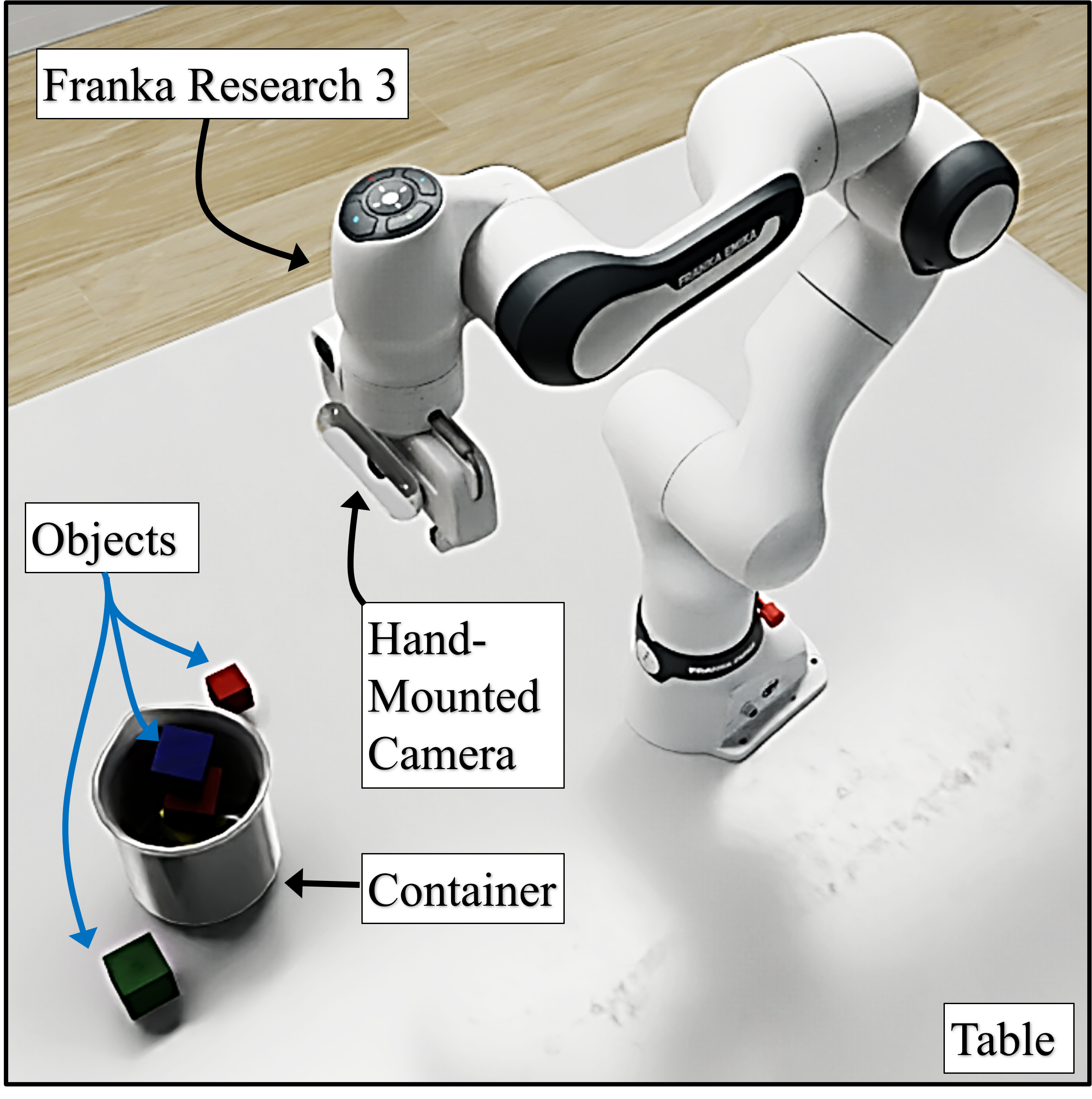}
    \vspace{-0.1cm}
    \caption{Simulated tabletop manipulation environment used for evaluation. The setup includes an FR3 robot arm, colored block objects, and an open container surrogate that can serve as a placement target or occluding object.}
    \label{fig:tabletop_setup}
\end{figure}

The system is evaluated in an NVIDIA Isaac Sim tabletop environment with a simulated Franka Research 3 robot, parallel-jaw gripper, colored manipulation objects, and a neutral open-container surrogate, as shown in Fig.~\ref{fig:tabletop_setup}. ROS~2 Jazzy and MoveIt~2 provide motion planning and execution. The language planner is accessed through an API-hosted GPT-5 mini model, while Qwen3-VL 8B provides limited local semantic verification. The local VLM is used only to confirm semantic roles, such as identifying an open-container surrogate or resolving visually similar candidates; it does not define coordinates, grasp poses, release locations, keepout dimensions, or task success.

\begin{figure}[t]
    \centering
    \includegraphics[width=0.9\linewidth]{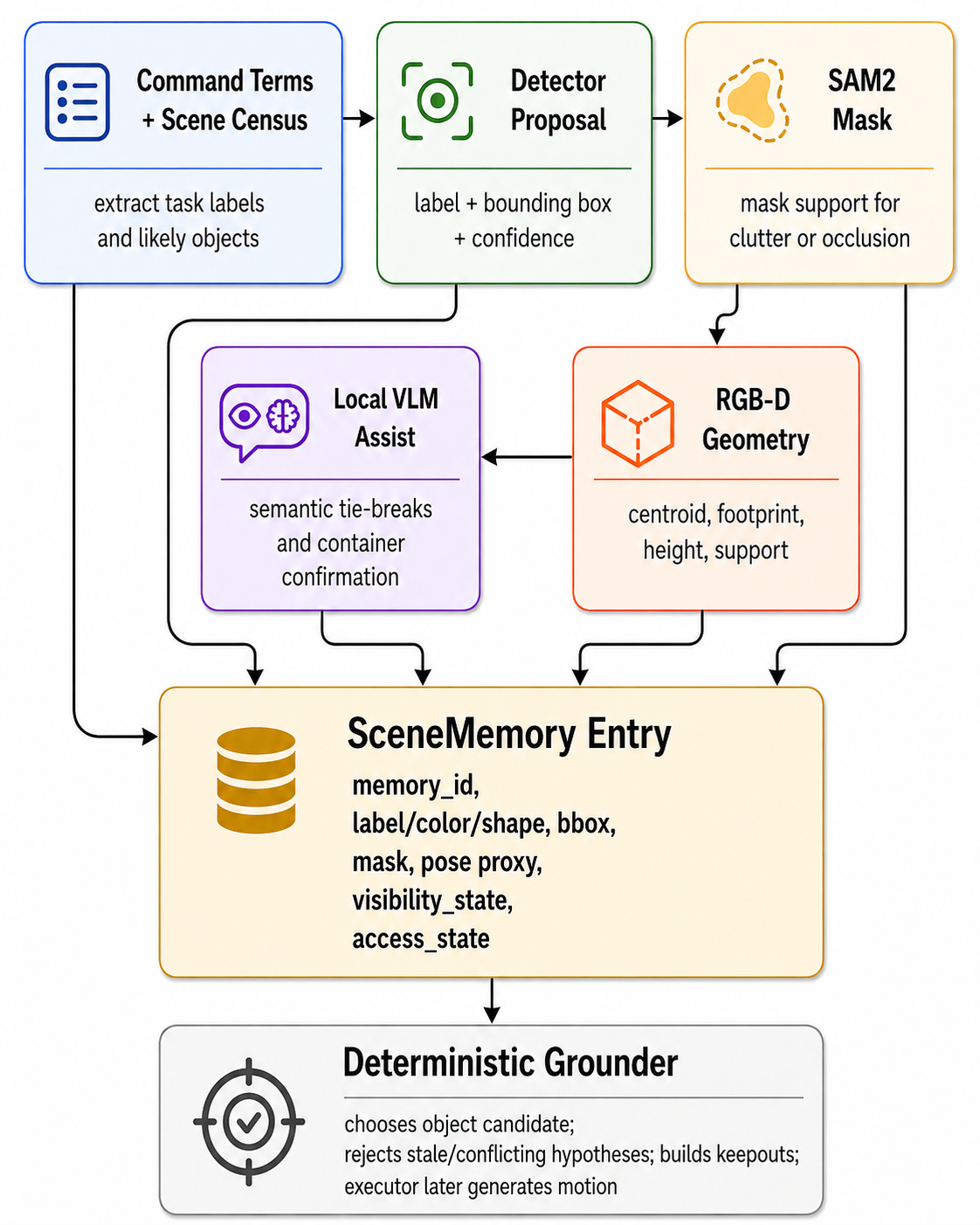}
    \caption{SceneMemory construction and deterministic candidate selection pipeline. Command terms and scene census guide open-vocabulary detection, SAM2 mask refinement, RGB-D geometric estimation, and limited local VLM assistance. These inputs are fused into a SceneMemory entry containing semantic, geometric, and accessibility attributes, which is then used by the deterministic grounder for object selection and keepout-aware grounding.}
    \label{fig:scene_memory_pipeline}
\end{figure}

\begin{figure}[t]
    \centering
    \includegraphics[width=0.7\columnwidth]{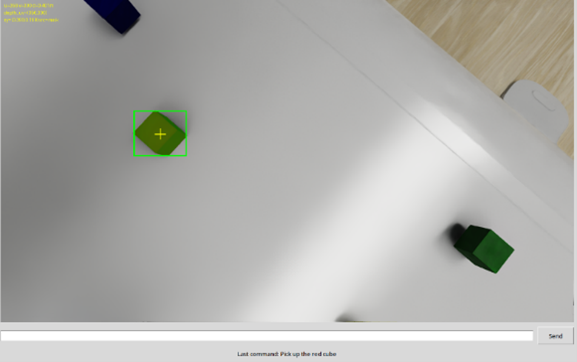}
    \caption{Example perception output during a commanded pickup task. The detector localizes the requested object in the RGB scene using a bounding box and center marker, providing the candidate-level visual evidence that is later fused with mask support, RGB-D geometry, and SceneMemory for deterministic grounding.}
    \label{fig:perception_example}
\end{figure}

SceneMemory connects perception, planning, grounding, and execution through stable memory identifiers. Each entry stores semantic labels, color, shape, bounding box, segmentation mask, depth-derived pose proxy, visibility, and access state. Access annotations indicate whether an object is top-down feasible, side feasible, occluded, keepout-sensitive, or requires re-observation, but final feasibility is determined by deterministic modules. SceneMemory is constructed from open-vocabulary detection, SAM2 mask refinement, RGB-D geometric estimation, and limited VLM assistance, as summarized in Fig.~\ref{fig:scene_memory_pipeline}. Task-relevant command terms and scene candidates are detected using open-vocabulary labels, bounding boxes, and confidence scores, as illustrated in Fig.~\ref{fig:perception_example}. Refined masks and RGB-D data are then used to estimate centroids, tabletop footprints, approximate heights, and support relationships, which are fused into SceneMemory entries for symbolic planning and deterministic grounding.

\subsection{Constrained Planning and Authority Boundary}
The planner is implemented as a prompted interface to a frozen pretrained language model, not as a robot controller. Its output is limited to symbolic fields, including task type, source and target references, placement relation, approach preference, avoid-object identifiers, re-observation flag, and an auditable reason. Metric outputs such as coordinates, joint angles, gripper widths, keepout dimensions, release heights, waypoints, or success labels are rejected before execution.

This boundary is central to the framework because it allows foundation models to contribute semantic reasoning without allowing hallucinated geometric quantities to directly control the robot. For example, the planner may indicate that a symbolic object should be avoided, but it may not define the keepout volume. Similarly, it may request a side approach or identify a target container surrogate, but the corresponding approach pose, insertion center, and release location are determined only by deterministic modules.

\subsection{Deterministic Grounding, Execution, and Feedback}
After schema validation, the deterministic grounder maps the symbolic plan to a predefined manipulation behavior and computes the metric parameters needed for execution. Object grounding prioritizes explicit command references and existing SceneMemory identifiers, then uses detector labels, color and shape attributes, mask support, and depth-derived geometry. Local VLM confirmation may serve only as a semantic tie-breaker and cannot override deterministic geometric checks.

Robot motion is executed through predefined routines, including pick, place, pick-and-place, re-observe, approach, retreat, grasp, lift, and release. Their metric parameters are computed at runtime from RGB-D geometry, object footprints, access annotations, workspace limits, and keepout constraints. The resulting Cartesian pose goals and gripper commands are sent to the motion interface, while MoveIt~2 performs collision checking, trajectory generation, and execution. Pickup requires verified approach, grasp contact, lift, and carry consistency, while placement requires geometric acceptance of the target region and post-placement visual confirmation of the commanded state.

Keepout handling and replanning follow the same bounded-authority structure. When the planner specifies an object to avoid, the executor retrieves its depth-derived footprint from SceneMemory, expands it using gripper- and task-dependent margins, and registers it as a collision object in the MoveIt~2 planning scene. Thus, the planner selects what should be avoided, while the executor determines the corresponding metric keepout. If contact, force, grasp, placement, or visual verification indicates failure, the executor updates SceneMemory and returns a symbolic failure description to the planner, which may request re-observation, revise the object reference, change the approach preference, or update avoid-object identifiers. Any revised plan must again satisfy the same schema constraints before grounding and execution.

\subsection{Evaluation Protocol}

The evaluation separates each trial into four observable stages: schema validation, deterministic grounding, collision-valid motion execution, and final outcome verification. The baseline followed the same command set, perception pipeline, SceneMemory representation, deterministic grounder, executor, MoveIt 2 interface, and scoring protocol, but disabled the local VLM assistance modules. A trial is counted as successful only when the final visual label confirms the commanded state change, and terminal text produced by the system is not treated as sufficient evidence of success. This structure supports failure localization by distinguishing whether an unsuccessful trial resulted from invalid symbolic output, lack of physically plausible grounding, motion-planning failure, or failed final visual verification. 

\begin{figure*}[t]
    \centering
    \includegraphics[width=\textwidth]{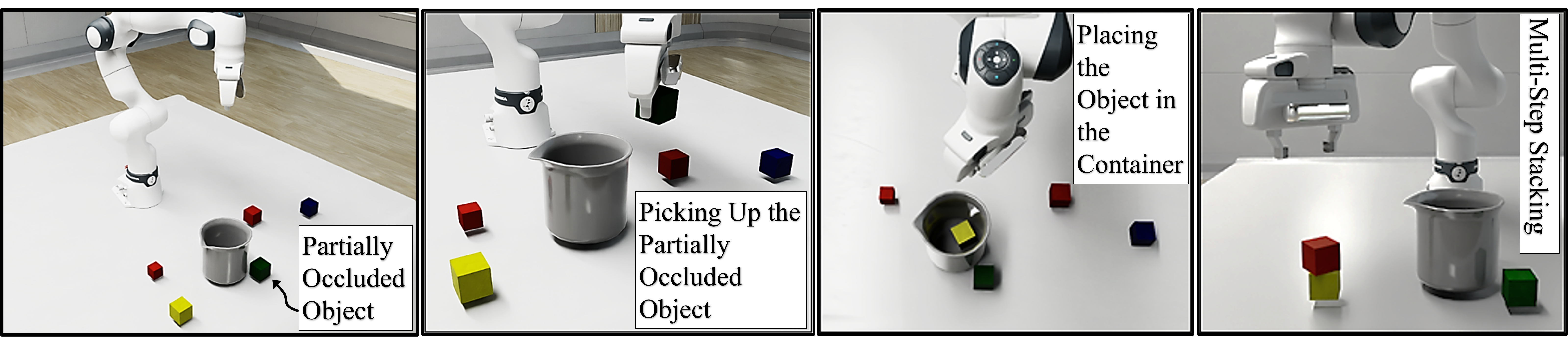}
    \caption{Representative Isaac Sim trial artifacts from the evaluated manipulation tasks, including partially occluded pickup, object pickup, container placement, and multi-step stacking.}
    \label{fig:results_sequence}
\end{figure*}

\section{RESULTS AND DISCUSSION}

The evaluation was conducted using a 40-trial artifact-complete Isaac Sim test matrix, with five command families and eight trials per family: clear-object pickup, size-distractor pickup, partially occluded pickup, container placement, and sequential manipulation. A trial was counted as successful only when the final visual state confirmed completion of the commanded task; terminal text alone was not used as a success criterion. Representative trial artifacts are shown in Fig.~\ref{fig:results_sequence}. The proposed VLM-assisted bounded-authority stack completed 37 of 40 trials, corresponding to an overall success rate of 92.5\%, while the baseline completed 4 of 40 trials, corresponding to 10.0\%. This baseline isolates the contribution of local VLM assistance while keeping the rest of the execution stack unchanged.

\begin{table}[t]
\centering
\caption{Evaluation results by command family.}
\label{tab:evaluation_matrix}
\begin{tabular}{lccc}
\hline
\textbf{Command family} & \textbf{N} & \textbf{Proposed stack} & \textbf{Baseline} \\
\hline
Clear-object pickup & 8 & 8/8 (100.0\%) & 2/8 (25.0\%) \\
Size-distractor pickup & 8 & 7/8 (87.5\%) & 2/8 (25.0\%) \\
Partially occluded pickup & 8 & 6/8 (75.0\%) & 0/8 (0.0\%) \\
Container placement & 8 & 8/8 (100.0\%) & 0/8 (0.0\%) \\
Sequential manipulation & 8 & 8/8 (100.0\%) & 0/8 (0.0\%) \\
\hline
Overall & 40 & 37/40 (92.5\%) & 4/40 (10.0\%) \\
\hline
\end{tabular}
\end{table}

The results support the central argument of this paper: foundation models are useful for semantic interpretation and task structuring, but physically consequential decisions should remain under deterministic robot-side authority. The proposed stack improved performance by using local VLM assistance for object identity, size ambiguity, container-surrogate recognition, access-state interpretation, and post-action visual verification. Importantly, these gains were achieved without allowing the LLM or VLM to generate robot coordinates, grasp poses, keepout geometry, gripper commands, trajectory waypoints, or success labels. This reinforces the motivation of the framework: the key challenge is not only whether a foundation model can infer user intent, but whether the architecture prevents ungrounded semantic outputs from becoming unsafe or invalid robot motion.

Performance trends across the command families further clarify the role of bounded semantic assistance. Clear-object pickup served as the basic manipulation control condition, requiring reliable object selection, grasp execution, and lift verification; the proposed stack completed all 8 trials, compared with 2 of 8 for the baseline. Size-distractor pickup introduced semantic ambiguity by placing a distractor with the same color and shape category as the target but with a different size. The proposed stack completed 7 of 8 trials, while the baseline again completed only 2 of 8, suggesting that VLM assistance improved semantic disambiguation while deterministic grounding preserved physical feasibility. Partially occluded pickup was more difficult because it required access-state reasoning, re-observation, and side-approach behavior. Although success decreased to 6 of 8, the baseline completed 0 of 8, indicating that the proposed framework can support occlusion-aware recovery without relaxing deterministic safety checks.

The container-placement and sequential-manipulation trials demonstrate the importance of SceneMemory and final visual verification beyond single-step object selection. Container placement required preserving target identity, validating the opening region, computing a safe release location, and confirming the post-placement state; the proposed stack completed all 8 trials, while the baseline completed none. Sequential manipulation added the need to maintain and update world state across multiple actions without resetting the scene. The proposed stack again completed all 8 trials, indicating that SceneMemory could be updated after each manipulation stage and reused for subsequent planning and execution. These results show that the framework functions as a traceable manipulation pipeline with explicit intermediate state representation, rather than as a one-shot object selector.

Compared with language-conditioned planning systems that rank skills or generate robot programs, the proposed framework enforces a stricter separation between symbolic task structure and metric execution authority. The LLM does not write robot code or produce motion parameters; it generates a constrained symbolic plan that must pass schema validation before grounding. Compared with VLA-style systems, the framework does not attempt end-to-end action generation from large-scale robot data. Its advantage is auditability: when a trial fails, the failure can be localized to object grounding, access-state interpretation, motion feasibility, release validation, or final visual verification. The baseline comparison also clarifies that the local VLM improves semantic robustness but does not replace deterministic grounding, motion planning, or outcome verification.

A limitation of this study is that the evidence is restricted to simulated FR3 tabletop manipulation in NVIDIA Isaac Sim. Physical FR3 deployment is not claimed as a completed result. The task set is also limited to clear-object pickup, size-distractor pickup, partially occluded pickup, container placement, and sequential manipulation supported by the current executor. In addition, the local VLM provides only bounded semantic assistance and does not own metric geometry, motion planning, grasp selection, release approval, or success labels. This improves auditability but may reduce flexibility compared with end-to-end VLA policies. The current SceneMemory representation also stores object hypotheses, approximate pose proxies, visibility states, and discrete access states; it is not a complete semantic SLAM map or a learned dynamics model.

Future work will transfer the framework to the physical FR3 platform and evaluate its reliability under real sensing noise, calibration error, contact uncertainty, lighting variation, and hardware latency. The evaluation should also be expanded to more diverse objects, container shapes, occlusion geometries, placement relations, and longer sequential tasks. Additional work is needed to improve SceneMemory with temporal tracking, uncertainty estimates, object-state histories, and re-identification after motion, occlusion, or contact. A larger benchmark should also report failure causes by layer, including perception, schema validation, grounding, motion planning, grasp verification, release checking, and final visual verification.

\section{CONCLUSION}

This paper presented a bounded-authority LLM--VLM framework for zero-shot robotic manipulation in a simulated FR3 tabletop environment. The framework allows foundation models to support symbolic planning and semantic verification, including object-reference interpretation, size-based disambiguation, container-surrogate confirmation, and post-action visual checking, while preventing them from generating metric robot commands such as coordinates, grasp poses, keepout dimensions, trajectory waypoints, gripper commands, or success labels. Instead, object grounding, pose selection, keepout construction, motion execution, grasp verification, release approval, and final task scoring remain under deterministic robot-side control. Across the 40-trial Isaac Sim evaluation, the proposed stack achieved a 92.5\% success rate compared with 10.0\% for the baseline with VLM assistance disabled, supporting the central claim that foundation models can improve semantic robustness when their authority is explicitly bounded. These results show that the proposed architecture provides a traceable middle ground between conventional deterministic manipulation pipelines and end-to-end foundation-model control, with future work focused on physical FR3 deployment, broader task and object diversity, improved SceneMemory tracking, and more detailed failure attribution across perception, grounding, execution, and verification stages.

\bibliographystyle{IEEEtran}
\bibliography{References}

\end{document}